\pdfoutput=1

\documentclass[11pt]{article}

\usepackage[]{acl}

\usepackage{times}
\usepackage{latexsym}

\usepackage[T1]{fontenc}

\usepackage[utf8]{inputenc}

\usepackage{microtype}

\usepackage{microtype}
\usepackage{graphicx}
\usepackage{subfigure}
\usepackage{booktabs} 
\usepackage{tabularx}
\usepackage{amsmath}
\usepackage{wrapfig}

\usepackage[textsize=tiny]{todonotes}
\usepackage{tikz}
\usepackage{pgfplots}
\usetikzlibrary{patterns}
\definecolor{MYBLUE}{RGB}{40, 120, 181}
\definecolor{MYLBLUE}{RGB}{154, 201, 219}
\definecolor{MYORANGE}{RGB}{248, 172, 140}
\definecolor{MYRED}{RGB}{200, 36, 35}
\definecolor{MYPINK}{RGB}{255, 136, 132}

\usepackage{amsmath}
\usepackage{amssymb}
\usepackage{mathtools}
\usepackage{amsthm}
\theoremstyle{plain}

\theoremstyle{definition}

\theoremstyle{remark}

\usepackage{amsmath,amsfonts,bm}

\def\eqref#1{equation~\ref{#1}}

\def\1{\bm{1}}

\def\vh{{\bm{h}}}

\def\vp{{\bm{p}}}

\DeclareMathAlphabet{\mathsfit}{\encodingdefault}{\sfdefault}{m}{sl}
\SetMathAlphabet{\mathsfit}{bold}{\encodingdefault}{\sfdefault}{bx}{n}

\newcommand{\E}{\mathbb{E}}

\newcommand{\R}{\mathbb{R}}

\newcommand{\softmax}{\mathrm{softmax}}

\usepackage{mathtools}
\usepackage{multirow}
\usepackage{multicol}
\usepackage{booktabs}
\usepackage{subfigure}

\DeclareMathOperator*{\kldiv}{KL}

\DeclareMathOperator*{\poscontrast}{P-Cf}

\title{PCL: Peer-Contrastive Learning with Diverse Augmentations for Unsupervised Sentence Embeddings}

\author{Qiyu Wu$^{1}$\thanks{\hspace{1.5mm}Work done during the internship at Microsoft.} ,
        Chongyang Tao$^{2}$, 
        Tao Shen$^{2}$,
        Can Xu$^{2}$,
        Xiubo Geng$^{2}$,
        Daxin Jiang$^{2}$\thanks{\hspace{1.5mm} Corresponding author.} \\
$^{1}$\normalfont{The University of Tokyo}, Tokyo, Japan \\
$^{2}$\normalfont{Microsoft Corporation} \\
$^{1}$\normalfont{\texttt{qiyuw@g.ecc.u-tokyo.ac.jp}} \\
$^{2}$\normalfont{\texttt{\{chotao,shentao,caxu,xigeng,djiang\}@microsoft.com}}
}

\begin{document}
\maketitle

\begin{abstract}
Learning sentence embeddings in an unsupervised manner is fundamental in natural language processing. Recent common practice is to couple pre-trained language models with unsupervised contrastive learning, whose success relies on augmenting a sentence with a semantically-close positive instance to construct contrastive pairs. Nonetheless, existing approaches usually depend on a mono-augmenting strategy, which causes learning shortcuts towards the augmenting biases and thus corrupts the quality of sentence embeddings. A straightforward solution is resorting to more diverse positives from a multi-augmenting strategy, while an open question remains about how to unsupervisedly learn from the diverse positives but with uneven augmenting qualities in the text field. As one answer, we propose a novel Peer-Contrastive Learning (PCL) with diverse augmentations. PCL constructs diverse contrastive positives and negatives at the group level for unsupervised sentence embeddings. PCL performs peer-positive contrast as well as peer-network cooperation, which offers an inherent anti-bias ability and an effective way to learn from diverse augmentations. Experiments on STS benchmarks verify the effectiveness of PCL against its competitors in unsupervised sentence embeddings.\footnote{Our implementation is available at \url{https://github.com/qiyuw/PeerCL}.}
\end{abstract}

\section{Introduction}
\label{sec:1}
Sentence embedding learning, which aims at deriving semantically meaningful fixed-sized vectors for sentences, is a natural language processing (NLP) technique of great significance, especially for time-sensitive downstream tasks~\cite{reimers2019sentence}. 
Recently, contrastive learning (CL) is proven effective to learn representation~\cite{Wu2018UnsupervisedFL,Tian2020ContrastiveMC,He2020MomentumCF} and substantially improve its performance~\cite{yan2021consert,gao2021simcse} when coupling with pre-trained language models (PLMs). 
The main idea of contrastive learning for sentence embedding is pulling semantic neighbors together and pushing semantic non-neighbors apart~\cite{hadsell2006dimensionality}, which naturally requires effective contrastive pairs. As effective contrastive pairs are usually scarce and require much human effort to collect, how to learn sentence embeddings in an fully unsupervised manner has become a challenging yet attractive research area~\cite{Wang2021TSDAEUT,Giorgi2021DeCLUTRDC}.
\begin{table}[t]
\centering
\resizebox{0.47\textwidth}{!}{
\begin{tabular}{cccc}
\toprule
\textbf{Augmenting}   & \textbf{Order}      & \textbf{N-gram} & \textbf{Bag-of-words} \\ \midrule
\emph{Shuffled Sentence} & $\times$ & $\times$     & $\checkmark$   \\ 
\emph{Inversed Sentence} & $\times$ & $\checkmark$      & $\checkmark$   \\ 
\emph{Word Repetition} & $\checkmark$ & $\times$      & $\checkmark$   \\ 
\emph{Word Deletion} & $\checkmark$ & $\times$     & $\times$   \\ \bottomrule
\end{tabular}
}
\caption{\small Text augmentation strategies change semantics in the sentence but still has shortcuts to learn. Employing limited number of strategies causes learning shortcuts towards the augmenting bias.}
\label{tab:bias}
\end{table}
\begin{figure*}[]
    \centering
    \subfigure[Conventional CL]{
        \includegraphics[width=.25\linewidth]{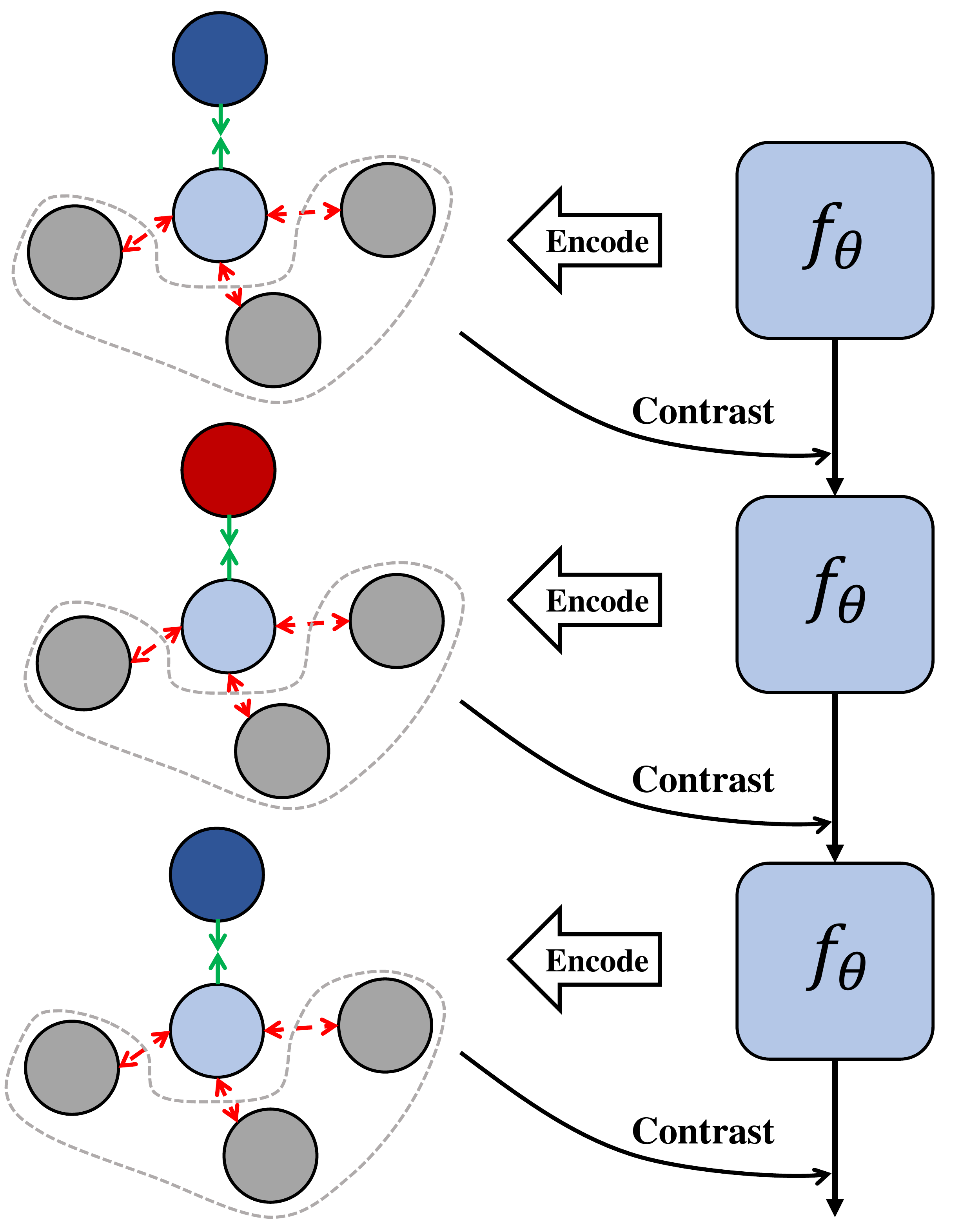}
        \label{fig:intro1}
    } \hspace{-2mm}
    \subfigure[PCL with diverse augmentations]{
        \includegraphics[width=.35\linewidth]{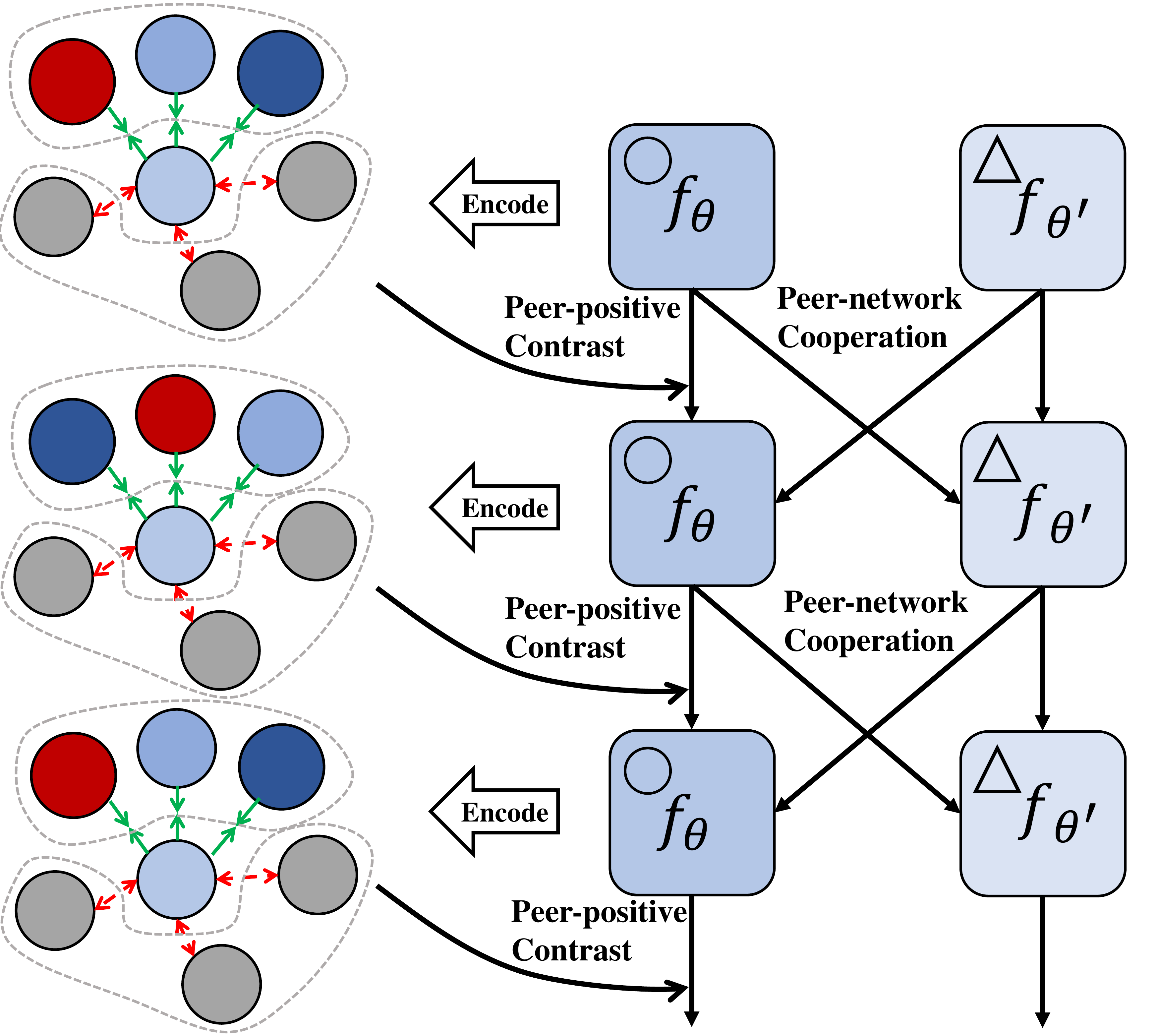}
        \label{fig:intro2}
    } \hspace{-2mm}
    \subfigure[Cooperative learning objective]{
        \includegraphics[width=.26\linewidth]{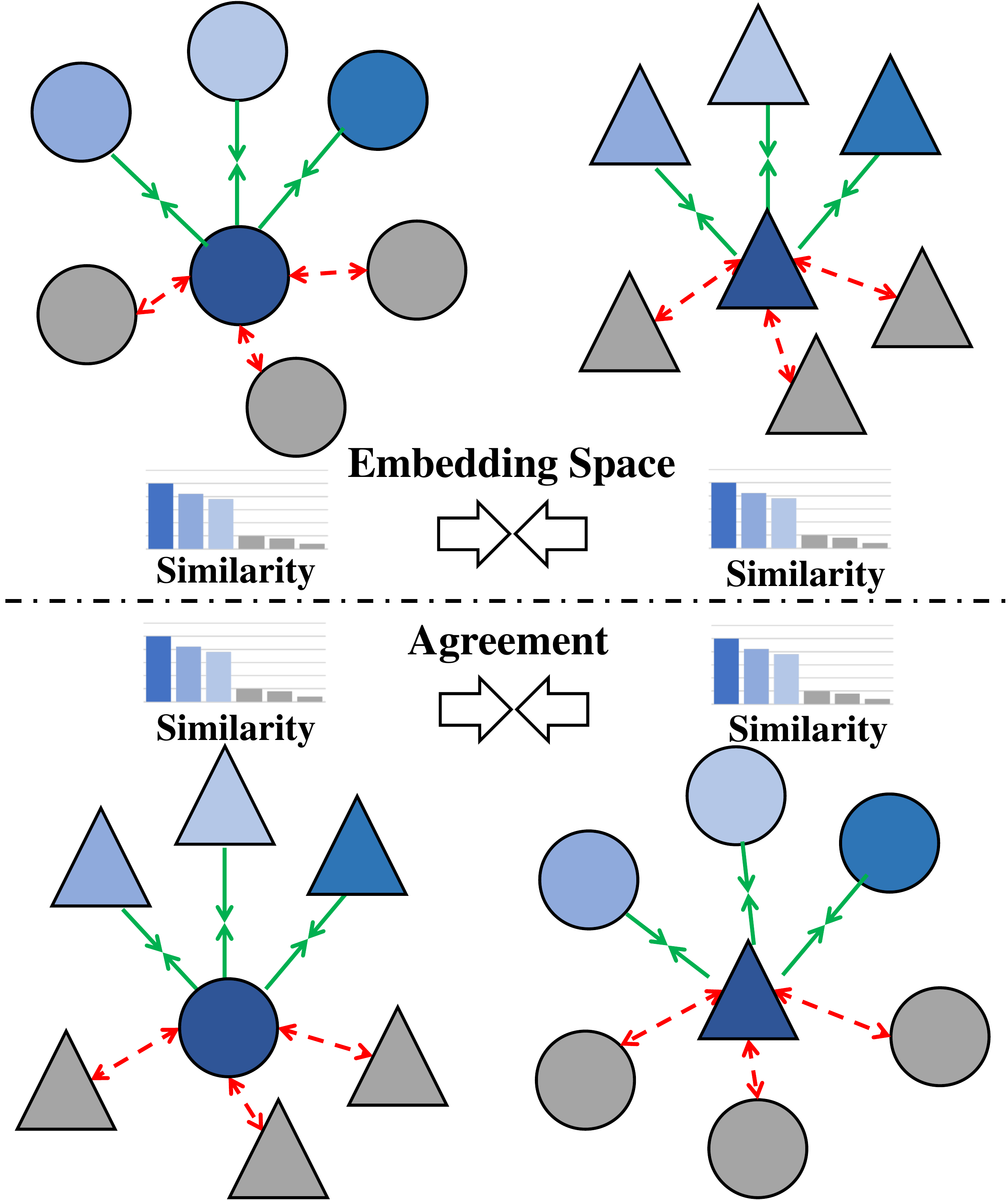}
        \label{fig:intro3}
    }
    \caption{
    \small
    The main idea of PCL. Circles and triangles denote contrastive instances encoded by $f_{\theta}$ and $f_{\theta'}$, respectively. The red ones denote ineffective positives, blue ones denote effective positives, and gray ones denote negatives.
    \textbf{(a)} Conventional CL on ineffective positive instances causes shortcut learning towards mono-augmenting biases.
    \textbf{(b)} PCL improves the probability of `at-least-one' effective positives and performs contrasts among peer positives and cooperation between peer networks. PCL maintains two networks, and each network learns from its peer network to achieve a common agreement.
    \textbf{(c)} Cooperative learning objective considers peer-positive contrasts to achieve PCL with diverse augmentations. The top panel illustrates the consistency of embedding space, while the bottom panel illustrates the agreement between the peer networks.
    }
    \label{fig:intro}
\end{figure*}

The key to unsupervised contrastive learning for sentence embedding is to augment a given anchor sentence with an effective positive instance to construct the pairs. Hence, many efforts have been made to design augmentation methods by adding noises or using heuristics, which mainly fall into two categories in terms of augmentation format -- \emph{discrete} and \emph{continuous}. 
The former operates directly on words or n-grams in the sentence, e.g., synonym substitution~\cite{su2021improving}, shuffling and word deletion~\cite{yan2021consert}. The latter operates on latent embeddings derived by neural encoder(s), e.g., SimCSE with twice dropouts~\cite{gao2021simcse}.
However, these existing approaches usually depend on a mono-augmenting format (i.e., either discrete or continuous) with a limited number of augmenting strategies, which suffer from learning shortcuts \cite{Ilyas2019AdversarialEA,Du2021TowardsIA} towards the augmenting biases and thus corrupt the quality of learned embeddings. 
For example, learning shortcuts caused by discrete augmenting biases are shown in Table~\ref{tab:bias}, and SimCSE based solely on dropout in continuous format is biased towards the sentence length~\cite{Wu2021ESimCSEES}.

To prevent the learning shortcuts caused by the potential biases from the mono-augmenting strategy, one straightforward solution coming into our mind is that we can consider more diverse augmentations for a given sentence in both continuous and discrete formats. 
Besides learning from diverse instances for inherent anti-bias ability, it can also bring a great opportunity for more effective learning. 
In particular, controlling the qualities of the noisy and heuristic augmentations for different sentences is almost impossible\footnote{It may cause poisonous positive. For example, given a sentence \textit{``A dog is chasing a cat.''}, a possible shuffled sentence is \textit{``A cat is chasing a dog.''}, which is semantically different.}. 
As illustrated in Figure~\ref{fig:intro1}, the resulting contrastive instances may become ineffective and even poisonous for conventional CL. 
Nonetheless, diverse instances from various augmentation strategies can notably improve the possibility of \emph{at-least-one} effective positives in the contrastive instances, so how to leverage the rich relations among the diverse augmentations for more effective CL is worth exploiting.

To this end, we propose a brand-new Peer-Contrastive Learning (PCL) with diverse augmentations, and an illustration of its overall framework is shown in Figure~\ref{fig:intro2}.
Firstly, PCL not only performs the vanilla positive-negative contrasts but also takes the opportunity to learn the rich structured relations among the diverse positives (i.e., \emph{peer-positive}) to highlight the more possibly effective ones. 
Then, to learn the structured relations in a fully unsupervised manner, we propose a cooperative learning framework consisting of 
two peer embedding networks (i.e., \emph{peer-network}). The two networks learn from each other to prevent error reinforcement in sole-network and achieve a common agreement from different views (as shown in Figure~\ref{fig:intro3}).
Consequently, the sentence embedding network is equipped with (i) anti-bias abilities by CL on the diverse augmentations and (ii) improved effectiveness by the unsupervised PCL, leading to a high quality of sentence embeddings. 

We conduct experiments on 7 standard semantic textual similarity (STS) tasks~\cite{agirre2012semeval,agirre2013sem,agirre2014semeval,agirre2015semeval,agirre2016semeval,cer2017semeval,marelli2014sick} to evaluate PCL. Results demonstrate that PCL significantly outperforms state of the art on 7 STS tasks. Typically, PCL achieves a 2.85\% improvement over the previous best approach in the averaged Spearman's correlation of 7 STS tasks in the BERT$_\text{base}$ setting.
PCL also outperforms previous approaches across different PLMs initialization and model sizes.
Moreover, ablation study and analysis show that the two proposed components, i.e., peer-positive contrasts and peer-network cooperation, are both capable of improving unsupervised sentence embedding learning.
\section{Peer-Contrastive Learning (PCL)}

This section begins with a formal definition of unsupervised sentence embeddings, followed by detailed formulations of our PCL with diverse augmentations (\S\ref{sec:cl_nlp} and \S\ref{sec:cc_peers}). Lastly, we will elaborate on our training and inference procedure for unsupervised sentence embeddings (\S\ref{sec:train_infer}). 

\paragraph{Unsupervised Sentence Embedding.} Given a sentence $x_i \sim X$, the target of {this task} is to learn a neural network $f_{\theta}$ (parameterized by $\theta$) without any human-labeled data. 
Then, the network can be applied to $x$ and derive a dense real-valued vector representation, i.e., $\vh_i = f_{\theta}(x_i)\in\R^d$. 
Consequently, $\vh_i$ can be used to represent the semantics of $x_i$ and fulfill downstream sentence-related tasks, e.g., semantic textual similarity. 
Thereby, this task depends on the designs of unsupervised (a.k.a self-supervised) objectives based on $X$ to learn $f_{\theta}$ effectively.


\subsection{Contrastive Representation Learning} \label{sec:cl_nlp}

The recent common practice of representation learning in an unsupervised manner is contrastive learning~\cite{Zhang2020AnUS,yan2021consert,gao2021simcse}, which aims to learn effective representations to pull similar instances together and push apart the other instances.
Thereby, compared to supervised contrastive learning that has already offered contrastive pairs, how to augment the given anchor (e.g., an image and sentence) with effective positive and negative instances to construct the contrastive pairs is critical in the unsupervised scenario. 
More recently, a simple contrastive learning framework (SimCLR) is proposed in visual representation learning~\cite{chen2020simple}, which constructs positive instances by different views (e.g., chop) of an image then learn to minimize the following InfoNCE loss.
\begin{align}  \small
   & L^{(\text{c})}_{\theta}(X, \delta; \theta) =  \\
    \notag - & \E_{x_i\sim X}\left[\log \dfrac{e^{s[f_{\theta}(x_i),
    f_{\theta}(\delta(x_i))]/\tau}}
                 {\sum_{ x_j \sim X \wedge j\ne i \vee \delta(x_i) } e^{s[f_{\theta}(x_i), f_{\theta}(x_j)]/\tau} } \right],
\end{align}  
where $\delta(\cdot)$ denotes using a different view of the image $x_i$ as the positive instance during visual contrastive learning, $x_j$ denotes negative instances against $x_i$ to construct contrastive pairs with $\delta(\cdot)$, 
and $s[\cdot, \cdot]$ denotes a similarity metric between two dense vectors. 
And $L^{(\text{c})}_{\theta}$ denotes this loss function is optimized w.r.t the subscript $\theta$. 
It is also noteworthy that $x_j\sim X$ is usually implemented by using other in-batch instances during mini-batch SGD (a.k.a in-batch negatives). 

However, when switching to unsupervised sentence embedding, augmenting an input sentence by a fully random chop or permutation may become very intractable.
This is because these operations are most likely to destroy the original sentence in both semantics and syntax and cause trivial positive augmentations. 
Hence, many research efforts have been made to design $\delta$ for effective positive augmentations in the NLP community. 
These efforts mainly fall into two categories in terms of augmentation format -- `discrete' and `continuous'. 
Discrete augmentation format denotes operating directly on the inputted sentence, where $\delta$ is defined as word deletion, shuffling~\cite{yan2021consert}, back translation~\cite{Xie2020UnsupervisedDA}, etc.
In contrast, continuous ones operate on hidden states or network parameters, where $\delta$ is defined as network twice dropout~\cite{gao2021simcse}, etc. 

Nonetheless, compared to visual contrastive learning that barely introduces new data distribution for the positive instances, such heuristic augmentation methods in the text field cause shortcut learning~\cite{Ilyas2019AdversarialEA,Du2021TowardsIA} -- each method exposes the learning procedure to potential biases towards the augmented instances and thus corrupts the quality of learned embeddings. 
Therefore, existing unsupervised contrastive sentence embedding works usually depend on the limited- or even mono-augmenting methods for their positive instances and inevitably suffer from the biases in the positive instances. 


\subsection{Contrast-Cooperation with Peers} \label{sec:cc_peers}

To prevent learning shortcuts caused by the potential biases from mono-augmenting strategy and exploit rich relations among diverse augmentations for more effective positives, we propose a brand-new contrastive learning method, called peer-contrastive learning (PCL). Besides the vanilla contrastive objective, it contains a novel `contrast-cooperation' learning mechanism, which we will detail in the following. 

\subsubsection{Multi-Augmenting Strategy} 
First, we adopt a multi-augmenting strategy for extensively diverse augmentations. 
Given a sentence $x_i\sim X$, it considers extensive augmentation methods from both continuous and discrete perspectives. This can be formally written as 
\begin{align}
    \Delta = \{\delta_k | \delta_k\in \Delta^{(c)}\cup \Delta^{(d)} \},
\end{align}
where $\Delta$ denotes a set of multiple augmentation methods from both the continuous $\Delta^{(c)}$ set and discrete $\Delta^{(d)}$ set, and $|\Delta|=K$.
Then, we can obtain diverse augmented positives by applying $\Delta$ to a sentence $x_i \sim X$, i.e., 
\begin{align}
    \hat X^i = \{ x^i_k=\delta_k(x_i) | \delta_k\in\Delta \}.
\end{align}

The contrastive sentence embedding based on diverse positives can mitigate the biases towards mono-augmenting strategy, but it comes with a double-edged sword. That is, the $\hat x^i_k \sim \hat X^i$ varies a lot with many factors (e.g., input sentence $x_i$ and augmentation method $\delta_k$), making it hard to control the quality of each $x^i_k$. 
To one extreme, one augmentation can become ineffective and even poisonous if its semantics is largely changed and thus corrupt the model.

\subsubsection{Contrast among Peer Positives } 
To effectively learn from the uneven qualities of the augmented positives, we propose a brand-new peer-contrastive learning framework that not only performs the vanilla positive-negative contrast but a positive-positive contrast. 
This is because our diverse augmentations provide a great opportunity to model rich structured relations among the positives and improve the probability of `at-least-one' effective positive in $\hat X^i$. And the positive-positive contrast can mimic `peer-competition' to highlight more likely effective positives but weaken the others' effects by suppressing them. 

Formally, we first derive a group-wise probability distribution by contrasting the anchor $x_i$ with both diverse positives $\hat X^i$ and in-batch negatives $x_j \sim X \wedge j\ne i$. That is,
\begin{align}
    &\vp^{\text{P-Cf}}_{\theta^1, \theta^2}(x_i) \coloneqq \poscontrast(x_i, \Delta^{(d)};\theta^1, \theta^2) = \\
    \notag &\softmax(\{s[f_{\theta^1}(x_i),f_{\theta^2}(\hat x^i_k)/\tau]\}_{\hat x^i_k \sim\hat X^i} + \\
    \notag &~~~~~~~~~~~~~~~~~\{s[f_{\theta^1}(x_i), f_{\theta^2}(x_j) /\tau]\}_{x_j \sim X \wedge j\ne i}),
\end{align}
where `+' here denotes a union of two sets. 
Identical to the vanilla contrastive sentence embedding \cite{gao2021simcse}, we also leverage a $\softmax$ normalization to fulfill peer-contrast among augmented positives. 
Please note we introduce $\theta^1$ and $\theta^2$ for clear deliveries in the remaining sections, and the two parameters here can be either tied (i.e., $\theta^1 = \theta^2 = \theta$) or not.
Although using the augmented positives to `compete' each peer sounds attractive for contrastive learning, one critical question remains about how to learn merely from effective positives and guide the positive-peer contrasts $\vp^{\text{P-Cf}}_{\theta^1, \theta^2}(x_i)$ in a fully unsupervised way. 

\subsubsection{Cooperation across Peer Networks}

We propose a cooperative learning framework to learn contrasts among the augmented positives. It contains two peer embedding networks, and the two networks learn from each other to prevent error reinforcement in sole-network and achieve a common agreement from different views.
Specifically, we first build a peer network $\theta'$ which acts like a momentum encoder~\cite{He2020MomentumCF} to cooperatively learn with $\theta$. 
Here, $\theta$ and $\theta'$ can be untied or even heterogeneous. 
Then, we present the loss of the momentum-like cooperative learning, which is a combination of two Kullback–Leibler divergence losses. That is
\begin{align}
    L^{(p)}_{\theta, \theta'}(X, \Delta; \theta, \theta')=& \label{eq:kl_loss}\\
    \notag \E_{x_i}[\kldiv[
    \vp^{\text{P-Cf}}_{\theta, \theta}&(x_i),
    \vp^{\text{P-Cf}}_{\theta', \theta'}(x_i)
    ] + \\
    \notag ~ \kldiv&[
    \vp^{\text{P-Cf}}_{\theta, \theta'}(x_i),
    \vp^{\text{P-Cf}}_{\theta', \theta}(x_i)
    ]].
\end{align}
We call this `momentum-like' since $\theta'$ is not strictly a history of $\theta$ for more different views but depends on the second $\kldiv$ term to prevent significant divergence from $\theta$. 
Meanwhile, the first $\kldiv$ term is to reach an agreement between the main network $\theta$ and its peer network $\theta'$. 
This `learning-from-agreement' paradigm, including mutual-distillation~\cite{zhang2018deep} and denosing-by-agreement~\cite{Wei2020CombatingNL}, is proven effective in improving performance and learning from label noises by prior supervised works. In contrast, we hold a distinct motivation that the implicit relations in a group of diverse positives and in-batch negatives are expected to unsupervisedly match each peer embedding network with another view (e.g., structures and parameters). 

\paragraph{Remark.} The meaning of `\textit{peer}' has two folds: (i) It denotes that we want to learn the rich structured relations among the diverse augmented positives to highlight the effective ones; (ii) It involves a cooperative learning framework based on peer networks for modeling positive-positive contrasts to achieve PCL with diverse augmentations. 


\subsection{Training and Inference} \label{sec:train_infer}

\paragraph{Training Objective.} We write the loss as a combination of (i) our proposed contrast-cooperation learning for both $\theta$ and $\theta'$ simultaneously to highlight more effective positives and (ii) vanilla contrastive learning that is applied to $\theta$ and $\theta'$ separately and based on our diverse augmentations $\Delta$ for their strong anti-bias initializations. That is,
\begin{align}
\label{eq:loss}
    &L^{(\text{PCL})} = L^{(p)}_{\theta, \theta'}(X, \Delta; \theta, \theta') + \\
    \notag &~~~~~~~~~~~~\beta \sum_{\delta_k\in\Delta} \left[ L^{(\text{c})}_{\theta}(X, \delta_k; \theta) + L^{(\text{c})}_{\theta}(X, \delta_k; \theta') \right],
\end{align}
where $\beta$ is a hyperparameter to control if the training inclines to vanilla CL for the anti-bias purpose. Hence, $\beta$ could be annealing to provide strong unbiased initializations for different views at the beginning and then focus on contrast-cooperation with peers for more effective learning. 

\paragraph{Inference.} Due to the symmetrical learning paradigm, we empirically found $\theta$ and $\theta'$ achieve comparable performance in our pilot experiments. Nonetheless, we only use the main embedding network $\theta$ rather than ensembles them to encode each sentence for fair comparisons with its competitors.

\section{Experiments} \label{sec:experiments}
\begin{table*}[!ht]
    \centering
    \resizebox{0.9\textwidth}{!}{
    \begin{tabular}{lcccccccc}
    \toprule
       \textbf{Model} & \textbf{STS12} & \textbf{STS13} & \textbf{STS14} & \textbf{STS15} & \textbf{STS16} & \textbf{STSb} & \textbf{SICK-R} & \textbf{Avg.} \\
    \midrule[1pt]
        GloVe embeddings (avg.) & 55.14 & 70.66 & 59.73 & 68.25 & 63.66 & 58.02 & 53.76 & 61.32 \\ \midrule[0.01em]
        BERT$_\text{base}$~(first-last avg.) & 39.70&	59.38&	49.67&	66.03&	66.19&	53.87&	62.06&	56.70\\ 
        BERT$_\text{base}$-flow & 58.40&	67.10&	60.85&	75.16&	71.22&	68.66&	64.47&	66.55 \\ 
        BERT$_\text{base}$-whitening & 57.83& 66.90 & 60.90 & 75.08& 71.31& 68.24& 63.73& 66.28\\ 
        IS-BERT$_\text{base}$ & 56.77 & 69.24 & 61.21 & 75.23 & 70.16 & 69.21 & 64.25 & 66.58 \\
        CT-BERT$_\text{base}$ &61.63 &76.80 &68.47 &77.50 &76.48 &74.31 &69.19 &72.05 \\
        ConSERT$_\text{base}$ & 64.64 & 78.49 & 69.07 & 79.72 & 75.95 & 73.97 & 67.31 & 72.74\\
        SG-OPT$_\text{base}$ & 66.84 & 80.13 & 71.23 & 81.56 & 77.17 & 77.23 & 68.16 & 74.62\\
        BERT$_\text{base}$-Mirror & 69.10 & 81.10 & 73.00 & 81.90 & 75.70 & 78.00 & 69.10 & 75.50\\
        SimCSE-BERT$_\text{base}$  & 68.40 &	82.41 &	74.38 &	80.91 & 78.56 & 76.85 & 72.23 & 76.25\\
        \textbf{PCL-BERT$_\text{base}^{\dagger}$} & \textbf{72.84} & \textbf{83.81} &	\textbf{76.52} & 83.06 & \textbf{79.32} & \textbf{80.01} & \textbf{73.38} & \textbf{78.42} \\
        ~~~~\textbf{-- Avg. of seeds$^{\dagger*}$} & 72.74 & 83.36 &	76.05 & \textbf{83.07} & 79.26 & 79.72 & 72.75 & 78.14 \\
        \midrule
        RoBERTa$_\text{base}$~(first-last avg.) &40.88& 58.74& 49.07& 65.63& 61.48 & 58.55& 61.63& 56.57  \\
        RoBERTa$_\text{base}$-whitening &46.99& 63.24& 57.23& 71.36& 68.99& 61.36& 62.91& 61.73 \\
        DeCLUTR-RoBERTa$_\text{base}$ & 52.41& 75.19 & 65.52& 77.12& 78.63& 72.41& 68.62& 69.99 \\
        RoBERTa$_\text{base}$-Mirror & 66.60 & \textbf{82.70} & 74.00 & 82.40 & 79.70 & 79.60 & 69.70 & 76.40 \\
        SimCSE-RoBERTa$_\text{base}$  & 70.16 &	81.77 &	73.24 &	81.36 & 80.65 & 80.22 & 68.56 & 76.57\\
        \textbf{PCL-RoBERTa$_\text{base}^{\dagger}$} &71.13&82.38&\textbf{75.40}&83.07&\textbf{81.98}&\textbf{81.63}&\textbf{69.72}&77.90 \\
         ~~~~\textbf{-- Avg. of seeds$^{\dagger*}$}& \textbf{71.54} & \textbf{82.70} &	75.38 & \textbf{83.31} & 81.64 & 81.61 & 69.19 & \textbf{77.91} \\
    \midrule[1pt]
        BERT$_\text{large}$-flow &65.20 &73.39 &69.42& 74.92& 77.63& 72.26& 62.50& 70.76 \\
        SG-OPT$_\text{large}$ & 67.02& 79.42& 70.38& 81.72& 76.35& 76.16& 70.20& 74.46 \\
        ConSERT$_\text{large}$ & 70.69 & 82.96& 74.13& 82.78& 76.66& 77.53& 70.37& 76.45 \\
        SimCSE-RoBERTa$_\text{large}$ &72.86 &83.99 &75.62 &84.77 &\textbf{81.80} &81.98 &71.26 &78.90 \\
        \textbf{PCL-RoBERTa$_\text{large}^{\dagger}$} &74.08&84.36&76.42&85.49&81.76&82.79&71.51&79.49 \\
        ~~~~\textbf{-- Avg. of seeds$^{\dagger*}$}& 73.76 & 84.59 &	76.81 & 85.37 & 81.66 & \textbf{82.89} & 70.33 & 79.34 \\
        \textbf{PCL-BERT$_\text{large}^{\dagger}$} &74.87&\textbf{86.11}&78.29&\textbf{85.65}&80.52&81.62&\textbf{73.94}&\textbf{80.14} \\        ~~~~\textbf{-- Avg. of seeds$^{\dagger*}$}& \textbf{74.89} & 85.88 &\textbf{78.33} & 85.30 & 80.13 & 81.39 & 73.66 & 79.94 \\
    \bottomrule
    \end{tabular}
    }
    \caption{
    \small
        The models' performance comparison on STS tasks.
        We report the Spearman's correlation $\rho$ (\%) on 7 STS datasets.
        We highlight the highest numbers among models with the same pre-trained encoder. $^{\dagger}$: Our models. $^*$: We also run our models five times with different random seeds and report the \textit{average} of these five results on each column as the final number.
    }
    \label{tab:main_sts}
\end{table*}


\subsection{Unsupervised Corpus and Benchmark}
\label{sec:data}
We train and evaluate our model in a fully unsupervised manner. Following \citet{gao2021simcse}, we train our model on $10^6$ randomly sampled sentences from Wikipedia English. 
We evaluate our model on the semantic textual similarity (STS) tasks without using any STS training data. We report results on 7 datasets, namely the STS benchmark (STSb)~\citep{cer2017semeval}
the SICK-Relatedness (SICK-R) dataset~\citep{marelli2014sick} and the
STS tasks 2012 - 2016~\citep{agirre2012semeval,agirre2013sem,agirre2014semeval,agirre2015semeval,agirre2016semeval} (STS12-STS16). These datasets provide a gold standard semantic similarity between 0 and 5 for each sentence pair, which include texts from various domains, and we obtain them from the SentEval toolkit~\citep{conneau2018senteval}.

\subsection{Implementation of PCL}
\paragraph{Augmentation Strategies}
In this paper we utilize five unsupervised augmentation strategies that are commonly adopted in previous works~\cite{Wei2019EDAED, yan2021consert, gao2021simcse}. Augmentations from discrete perspectives $\Delta^{(d)}$ includes:
1) Shuffled Sentence (SS) shuffles the position of words in the sentence;
2) Inverted Sentence (IS) inverts the original sentence as the augmented sample;
3) Words Repetition (WR) duplicates part of words and randomly insert them into the original sentences;
4) Words Deletion (WD) deletes part of words in the sentences.
The augmentations from the continuous perspective $\Delta^{(c)}$ include Dropout (DP). It generates augmentation instances in the embedding level by passing the original sentence again into the encoder with different dropout masks.
More implementation details about augmentation are presented in \S~\ref{ap:implementation} due to the page limit.

\paragraph{Network Implementation}
We initialize the networks $\theta$ and $\theta^{'}$ with the PLMs checkpoint downloaded from Huggingface's \texttt{Transformers}\footnote{https://github.com/huggingface/transformers} of BERT~\cite{Devlin2019BERTPO} or RoBERTa~\cite{Liu2019RoBERTaAR}. The encoder consists of 12 and 24 Transformer layers for the base and large model, respectively. The hidden size is set to 768 and 1024, and the number of attention heads is set to 12 and 16 for base and large models, respectively. We choose the representation of the \texttt{[CLS]} token as the embedding of the input sentence. The hyperparameter $\beta$ is set to 1 for training simplicity without tuning, and the number of augmentations $K$ is 9 for base models.
Due to the computation resource limitation, particularly for large models, we set $K$ to 4, and the two networks $\theta$ and $\theta^{'}$ are tied, in which the cooperative learning is performed between the two passes through the network.

\paragraph{Training Setups.}
We follow common practices and carry out preliminary grid search on the development set of STSb to decide the hyper-parameter configuration. The learning rate is set to 3e-5 for base models and 1e-5 for large models, respectively. Except for learning rate, We use the same training hyper-parameters for all experiments with the batch size of 64 and the maximum length of 32. The temperature parameter $\tau$ is set to 0.05, and the dropout probability is set to 0.1. We train our model for 1 epoch and evaluate the model on the STSb development set every 125 steps, and keep the best checkpoint by following \citet{gao2021simcse}.

\paragraph{Evaluation Setups.}
We evaluate PCL on 7 STS tasks, including STS12-STS16, STSb, and SICK-R as introduced in \S~\ref{sec:data}. No training data of STS tasks are used during training and evaluation. \citet{gao2021simcse} has studied the evaluation settings for sentence embedding. We adopt their suggestions and follow the standard settings in Sentence-BERT~\cite{reimers2019sentence}. Specifically, we do not train an additional regressor for STSb and SICK-R, use Spearman's correlation as the metric, concatenate all tasks and report the overall Spearman's correlation. To fairly compare with previous approaches, we use the evaluation scripts released by \citet{gao2021simcse}\footnote{https://github.com/princeton-nlp/SimCSE}. Moreover, we train our model for five times with different random seeds and report the \textit{average} of these five results.
We also evaluate PCL on 7 transfer tasks~\cite{conneau2018senteval}. PCL achieves competitive performance and the detailed results are presented in Appendix~\ref{ap:transfer}. As mentioned in previous works~\cite{reimers2019sentence,gao2021simcse}, the main goal of sentence embeddings is to cluster semantically similar sentences. Hence we only take STS as the main results in this paper.


\subsection{Competitive Baselines}
We compare our model with previous state-of-the-art unsupervised sentence embedding approaches, including basic embedding approaches (e.g. average of GloVe~\cite{pennington2014glove}, BERT~\cite{Devlin2019BERTPO} or RoBERTa~\cite{Liu2019RoBERTaAR} embeddings) and contemporary contrastive learning approaches (e.g. IS-BERT~\cite{Zhang2020AnUS}, ConSERT~\cite{yan2021consert}, SG-OPT~\cite{Kim2021SelfGuidedCL}, Contrastive Tension~\cite{Carlsson2021SemanticRW}, DeCLUTR~\cite{Giorgi2021DeCLUTRDC}, Mirror-BERT~\cite{liu2021fast} and SimCSE~\cite{gao2021simcse}).
SGPT~\cite{muennighoff2022sgpt} and Sentence-T5~\cite{ni2021sentence} are proposed with new paradigm and far larger models, which underperform with comparable model size. Trans-Encoder~\cite{liu2021trans} proposes a cooperative method with in-domain pairwise data for mutual benefits of bi- and cross-encoder, making the results incomparable.
Please refer to \S~\ref{ap:baselines} for more details.

\subsection{Main Quantitative Results}
Experimental results on STS tasks are shown in Table~\ref{tab:main_sts}. We can find that our PCL significantly outperforms the previous best result on all seven tasks as well as the average STS score with a large margin compared to the baseline methods based on BERT$_{\text{base}}$ or RoBERTa$_{\text{base}}$ PLMs. Specifically, PCL improves the previous best result on average STS score from $76.25$ to $78.42$ for BERT$_{\text{base}}$ and $76.57$ to $77.91$ for RoBERTa$_{\text{base}}$, respectively. 
As SimCSE~\cite{gao2021simcse} did not report their performance on BERT$_{\text{large}}$, we compare all large models together, and the results are shown in the last rows of the table. We can observe that PCL outperforms the best result on all tasks apart from the STS16. Despite this, our PCL still obtain an improvement from $78.90$ to $80.14$ on the average STS score. 
Our PCL achieves more significant improvement over the base models than the large models,
And even so, PCL still outperforms SimCSE on almost all tasks in large models and all tasks in base models, which shows that PCL is effective across different model sizes and different types of PLMs. 

\subsection{Analysis of Diverse Augmentations}
\label{sec:ablation_augs}
Diversity and the number of augmentations are two crucial factors of PCL. In this section, we test the performance of PCL with varying $K$ and diversity. For PCL and all variants of PCL in this section, we train them for 5 times with different random seeds, and take the average as the final results.

\begin{figure}[!t]
    \centering
    \begin{tikzpicture}
    \begin{axis}[width=0.8\linewidth,height=.55\linewidth, ymin=76.3 , xtick={1,3,5,7,9}, ytick={76.5,77,77.5,78}, xlabel=Number of augmentations, ylabel=Avg. Spearman's correlation, ylabel near ticks, xlabel near ticks, font=\fontsize{7}{7}\selectfont, grid=major, xlabel shift={0}, ylabel shift={0}, nodes near coords, nodes near coords style={above},]
        \addplot[mark=*,  mark options={scale=1.3, fill=MYBLUE}, draw=MYBLUE][error bars/.cd,y dir=both, y explicit]
        coordinates {
        (1,77.06) +- (0.19,-0.19)
        (3,77.12) +- (0.21,-0.21)
        (5,77.59) +- (0.12,-0.12)
        (7,77.75) +- (0.29,-0.29)
        (9,78.14) +- (0.06,-0.06)
        };
    \end{axis}
\end{tikzpicture}
    \caption{\small Effect of the number of augmentations.}
    \label{fig:ablation_k}
\end{figure}
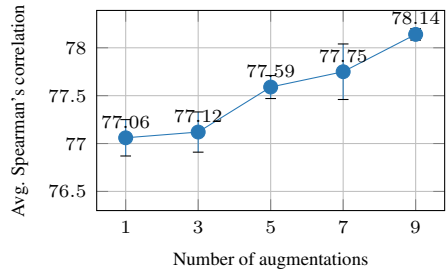

\paragraph{The number of augmentations.}
To mitigate the model bias towards the mono-augmenting strategy, we propose to augment the input sentence with a group of positive instances. The number of augmentations $K$ is a crucial hyper-parameter in this framework. To check if the performance of PCL is sensitive to $K$, we conduct experiments on PCL-BERT$_{\text{base}}$ with varying $K$ on 7 STS tasks and report the average STS score. We keep the diversity of augmentations $\Delta$ as much as possible when $K>1$.
As shown in Figure~\ref{fig:ablation_k}, the performance of PCL maintains an upward trend with increasing $K$.
This indicates that multiple augmentation strategy improves unsupervised sentence embeddings compared with learning with mono-augmenting strategy. This supports our motivation that contrastive learning with mono-augmenting strategy causes learning shortcuts. More detailed results on all 7 STS tasks are presented in \S~\ref{ap:diverse}.

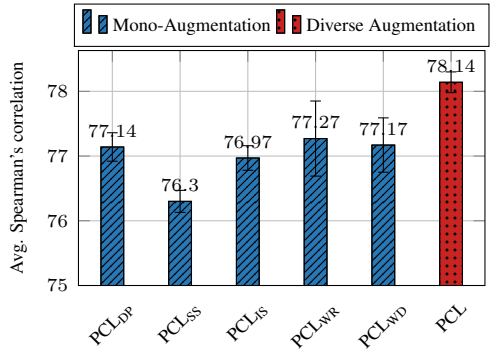
\begin{figure}[!t]
    \centering
    \begin{tikzpicture}
    \begin{axis}[
        width=0.9\linewidth,
        height=.61\linewidth,
        ybar,
        ymin=75,
        enlarge x limits =0.1,
        legend pos=north west,
        bar width=0.3cm,
        bar shift=0pt,
        ylabel=Avg. Spearman's correlation,
        ylabel near ticks,
        xlabel near ticks,
        xtick align=inside,
        xticklabel style={rotate=45},
        font=\fontsize{7}{7}\selectfont, legend columns=3,
        symbolic x coords={
            PCL$_\text{DP}$, PCL$_\text{SS}$, PCL$_\text{IS}$, PCL$_\text{WR}$, PCL$_\text{WD}$, PCL
        },
        xtick={PCL$_\text{DP}$, PCL$_\text{SS}$, PCL$_\text{IS}$, PCL$_\text{WR}$, PCL$_\text{WD}$, PCL},
        ytick={75,76,77,78},
        legend style={
                at={(0.5, 1.2)},
                anchor=north
            },
        nodes near coords,
        nodes near coords style={above},
        point meta=y,
        grid=major,
    ]
    \addplot[draw=black, fill=MYBLUE, postaction={pattern=north east lines}][error bars/.cd,y dir=both, y explicit] coordinates {
        (PCL$_\text{DP}$, 77.14) +- (0.22,-0.22)
        (PCL$_\text{SS}$, 76.30) +- (0.17,-0.17) 
        (PCL$_\text{IS}$, 76.97) +- (0.19,-0.19)
        (PCL$_\text{WR}$, 77.27) +- (0.58,-0.58)
        (PCL$_\text{WD}$, 77.17) +- (0.42,-0.42)
        };
    \addplot[draw=black, fill=MYRED, postaction={
         pattern=dots}][error bars/.cd,y dir=both, y explicit] coordinates {
        (PCL, 78.14) +- (0.16,-0.16)
        };
    \addlegendentry{Mono-Augmentation}
    \addlegendentry{Diverse Augmentation}
    \end{axis}
\end{tikzpicture}
    \caption{\small Effect of the diversity of augmentations.}
    \label{fig:ablation_diversity}
\end{figure}

\paragraph{The diversity of augmentations.}
Another critical factor is the diversity of augmentations. We fix $K=9$ and reduce the diversity of augmentations $\Delta$ to check if PCL is sensitive to the diversity. We conduct experiments on PCL-BERT$_{\text{base}}$ with $K=9$ but only use \emph{one} type of augmentation strategy from discrete and continuous perspective, respectively.
In other words, we keep at least one DP augmentation for all variants. We compare PCL with five mono-augmentation variants that are denoted as PCL$_{\text{DP}}$, PCL$_{\text{SS}}$, PCL$_{\text{IS}}$, PCL$_{\text{WR}}$ and PCL$_{\text{WD}}$, respectively. The details of augmentation strategies are introduced in \S~\ref{ap:implementation}.
Average Spearman's correlation scores of 7 STS tasks are shown in the Figure~\ref{fig:ablation_diversity}. 
Experimental results show that PCL significantly outperforms its mono-augmenting variants, even keeping the $K$ constant, indicating a better generalization. This supports our motivation that PCL with diverse augmentations can mitigate the shortcut learning biased towards mono-augmenting strategy.
Particularly, SimCSE utilizes dual-dropout to construct the contrastive pairs, hence the PCL$_{\text{DP}}$ variant (9 positive instances generated by the dropout) can be regarded as \textit{SimCSE w/ 9 augmented positive samples}. Our proposed contrasts and cooperation among peers improve SimCSE from 76.25 to 77.14, but the score is still lower than PCL with a large margin. This is another piece of evidence that shows the advantage of the diversity of augmentations. The detailed results on all 7 STS tasks are presented in \S~\ref{ap:diverse}.


\subsection{Ablation Study}
\label{sec:ablation}
We first check the impact of the proposed two components of PCL, peer-network cooperation and peer-positive contrast, i.e., the two terms in Equation~\ref{eq:loss} respectively. We designed two variants of PCL on PCL-BERT$_\text{base}$ by removing the cooperation loss and contrast loss, which are denoted as PCL$_\text{noP}$ and PCL$_\text{noC}$ respectively. To ensure the networks have the essential ability to learn sentence embeddings, we keep the contrastive loss with a single DP augmentation for PCL$_\text{noC}$, which is equal to the setting in Figure~\ref{fig:ablation_k} when $K=1$. We train each variant for five times with different random seeds and take the average of these seeds as the final results. As in Table~\ref{tab:ablation}, the average scores of PCL drop by $1.02$ and $1.08$ when removing the cooperation loss and contrast loss, respectively.
This indicates that our proposed peer cooperation and peer contrast are both beneficial to unsupervised learning of sentence embeddings. Among the two components, the peer cooperation loss plays a more important role as it incorporates contrasts among peer positives and peer networks, enabling an inherent anti-bias ability and an effective way to learn from diverse augmentations.
\newcolumntype{Y}{>{\centering\arraybackslash}X}
\begin{table}[!t]
\centering
\small
\begin{tabularx}{0.99\linewidth}{cYYY|Y}
\toprule
Tasks  & PCL           & PCL$_\text{noP}$   & PCL$_\text{noC}$ &SimCSE \\ \midrule
STS12  & \textbf{72.74} & 71.15          & 72.58   & 68.40         \\
STS13  & \textbf{83.36}          & 83.07 & 80.62    & 82.41         \\
STS14  & \textbf{76.05}          & 75.72 & 74.15    & 74.38         \\
STS15  & \textbf{83.07} & 82.93          & 82.31    & 80.91         \\
STS16  & \textbf{79.26}          & 78.37 & 79.23    & 78.56         \\
STSb   & \textbf{79.72} & 78.67          & 78.47    & 76.85        \\
SICK-R & \textbf{72.75} & 70.37          & 72.06    & 72.23        \\ \midrule
Avg.   & \textbf{78.14} & 77.12          & 77.06    & 76.25        \\ \bottomrule
\end{tabularx}
\caption{\small Ablation study on PCL-BERT$_\text{base}$. PCL$_\text{noP}$ denotes PCL w/o peer-network cooperation. PCL$_\text{noC}$ denotes PCL w/o peer-positive contrast.}
\label{tab:ablation}
\end{table}

\paragraph{Fixed peer encoder vs. trainable peer encoder.}
Particularly, We are also curious about the impact of `learning from agreement' (i.e., the second term in Equation~\ref{eq:kl_loss}) in the cooperative learning objective.
Therefore, we further test additional variants of PCL with a fixed peer encoder, denoted as PCL$_\text{FixedP}$. Specifically, we download a checkpoint of SimCSE as the peer encoder but fix its parameters while training.
Experimental results show that the performance of PCL$_\text{FixedP}$ drops with a large margin on STS12, STS13, STS14, STS15, STSb, and the average STS. The reason may be that although SimCSE is by far the best practice of sentence embeddings, it is still biased towards a mono-augmenting strategy.
Hence, cooperative learning with a biased peer network can be harmful to the network with diverse augmentations. This also indicates that it is necessary to simultaneously update the two peer networks and learn the agreement between them in PCL. Furthermore, there can be a considerable discrepancy between the embedding spaces produced by two methods, which hinders the cooperative training of two networks.
\begin{figure}[!t]
    \centering
    \begin{tikzpicture}
    \begin{axis}[
        width=\linewidth,
        height=.685\linewidth,
        ybar,
        enlarge x limits =0.1,
        legend pos=north west,
        bar width=0.2cm,
        ylabel=Spearman's correlation,
        ylabel near ticks,
        xlabel near ticks,
        xticklabel style={rotate=45},
        xtick align=inside,
        font=\fontsize{7}{7}\selectfont, legend columns=3,
        legend style={
                at={(0.75, 0.97)},
                anchor=north
            },
        symbolic x coords={
            STS12, STS13, STS14, STS15, STS16, STSb, SICK-R, Avg.
        },
        xtick={STS12, STS13, STS14, STS15, STS16, STSb, SICK-R, Avg.},
        ytick={70,72.5,75,77.5,80,82.5},
        point meta=y,
        grid=major,
    ]
    \addplot[draw=black, fill=MYBLUE, postaction={pattern=north east lines}][error bars/.cd,y dir=both, y explicit] coordinates {
        (STS12, 70.65) +- (1.02,-1.02)
        (STS13, 77.49) +- (0.32,-0.32) 
        (STS14, 71.69) +- (0.48,-0.48)
        (STS15, 79.45) +- (0.35,-0.35)
        (STS16, 78.87) +- (0.14,-0.14)
        (STSb, 76.55) +- (0.50,-0.50)
        (SICK-R, 72.46) +- (0.09,-0.09)
        (Avg., 75.31) +- (0.23,-0.23)
        };
    \addplot[draw=black, fill=MYRED, postaction={pattern=dots}][error bars/.cd,y dir=both, y explicit] coordinates {
        (STS12, 72.74) +- (0.58,-0.58)
        (STS13, 83.36) +- (0.40,-0.40) 
        (STS14, 76.05) +- (0.39,-0.39)
        (STS15, 83.07) +- (0.40,-0.40)
        (STS16, 79.26) +- (0.15,-0.15)
        (STSb, 79.72) +- (0.53,-0.53)
        (SICK-R, 72.75) +- (0.50,-0.50)
        (Avg., 78.14) +- (0.16,-0.16)
        };
    \addlegendentry{PCL$_\text{FixP}$}
    \addlegendentry{PCL}
    \end{axis}
\end{tikzpicture}
    \caption{\small Ablation study on whether updating the extra peer encoder. PCL$_\text{FixedP}$ denotes a variant of PCL that cooperatively learning with a fixed peer network.}
    \label{fig:ablation_fix}
\end{figure}
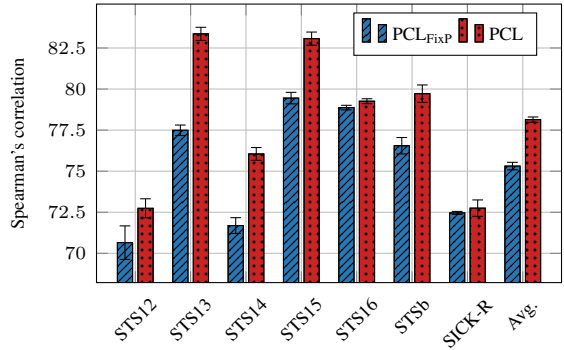

\section{Related Works}
\paragraph{Unsupervised Sentence Embedding.}
\label{sec:related-use}
Common practice of unsupervised sentence embedding is to take the average of pre-trained word embeddings~\cite{Mikolov2013EfficientEO,pennington2014glove} PLMs, like BERT~\cite{Devlin2019BERTPO} or RoBERTa~\cite{Liu2019RoBERTaAR}, \citet{wu2021taking} takes the average of word embeddings as context embedding to enhance the language pre-training. Other works also take the \texttt{[CLS]} embedding from the last layer of PLMs with post-processing~\cite{Li2020OnTS,su2021whitening}. Some works~\cite{skipthought,logeswaran2018an,hill-etal-2016-learning} directly train a deep model for sentence embeddings using co-occurrence information. Recent approaches couple PLMs with CL~\cite{Zhang2020AnUS,yan2021consert,Kim2021SelfGuidedCL,Carlsson2021SemanticRW,Giorgi2021DeCLUTRDC,gao2021simcse,xie2022stable} with a particular single strategy to construct contrastive pairs.
It is straightforward to extend mono-augmentation into multi-augmentation to learn expressive representations. For example, CLEAR~\cite{wu2020clear} uses various token/span manipulations for noise-invariant representations while Mirror-BERT~\cite{liu2021fast} employs several fast augmentation strategies. However, these methods usually take the augmented positives equally, regardless of the \textbf{uncontrollable qualities}. 
Thereby, we take a step further to consider the contrasts among the augmented positives to figure out which augmentation is relatively reasonable. This is achieved by our novel cooperative learning method with peer networks.
More related to our work, ESimCSE~\cite{Wu2021ESimCSEES} found learning on dual-dropout causes sentence length bias so it employs another augmentation strategy, i.e., word repetition, to prevent the length bias. However, word repetition introduces \textbf{learning shortcut} by itself, not to mention it makes the sentence unnatural and even \textbf{semantics-wrong}. 
To circumvent this dilemma, we propose exhaustive augmentations to ensure ``\textit{at-least-one}'' true positive and reduce learning shortcuts by complementary augmenting strategies. By doing so, PCL achieve a better performance on STS tasks.
Please refer to \S~\ref{ap:realted} for more discussion details.

\paragraph{Contrastive Learning.}
The main idea of CL is to pull semantic close neighbors close and push non-neighbors apart~\cite{hadsell2006dimensionality,pmlrzbontar21a}. It is shown to be a successful way to learn representation. Approaches in computer vision (CV)~\cite{chen2017sampling,Wu2018UnsupervisedFL,Tian2020ContrastiveMC,He2020MomentumCF,pmlrzbontar21a} try to make an image to be invariant to transformations on itself, while remaining discriminative to other images. More references in CV are discussed in the recent survey~\cite{jaiswal2021survey}. CL is also coupled with PLMs to learn sentence embeddings. But, recent works~\cite{Xiao2021WhatSN} argue that learning invariance to particular transformations may be harmful to the robustness of the model. This also supports our idea of leveraging diverse augmentations to improve unsupervised sentence embeddings from another angle.

\paragraph{Learning from Agreement.}
Another line of work close to ours is learning from agreement, e.g., Decoupling~\cite{malach2017decoupling}, Co-teaching~\cite{han2018co,yu2019does}, and mutual CL \citet{Yang2021MutualCL}. This paradigm has been proven effective in improving model performance and learning with label noises by prior fully-supervised works.
As text data is discrete and compositional, qualities of multiple augmentations can be uneven, which may corrupt the generalization of sentence embeddings. Besides widely used regularization like dropout~\cite{srivastava2014dropout} and weight decay~\cite{krogh1991simple}, we consider learning from agreement paradigm to offer a robust way to learn from our diverse positives.

\section{Conclusion}


In this paper, we propose a brand-new contrastive learning framework, dubbed as peer-contrastive learning (PCL), to capture rich relations among diverse positive peers and highlight effective positives, which are learned by cooperative learning by peer networks. Besides inherent anti-bias ability by diverse augmentations, it can learn from unsupervised corpus more effectively than vanilla contrastive in the text field. 
Experiments show that the number and diversity of augmentations are crucial to PCL. Ablation study also shows that the two components of PCL, i.e., peer-positive contrast and peer-network cooperation, are both beneficial to unsupervised CL for sentence embeddings.

\paragraph{Limitation.}
We also recognize that our PCL framework has its certain limitations:
(i) Due to peer positives and encoders, our framework needs higher (\textasciitilde3$\times$, i.e., 2.5
GPU-hours for base models) computation overheads compared to vanilla CL. Nonetheless, the acceptable extra overhead and same inference makes our framework still practical and scalable.
(ii) As a work addressing the general shortcut learning problem in a fundamental task, the proposed PCL is only evaluated on resources in English. It can be further extend to more applications such as in low-resource or other languages.
(iii) The performance of our framework relies on the choice of augmentation methods, and it is hard to strictly claim which combination of the methods is optimal except experimental verification. Although we have analysed the effect of varying combinations of augmentations with extensive experiments, we can only select several widely-adopted augmentations to evaluate the general effectiveness of our framework.

\section*{Acknowledgement}
We thank Yoshimasa Tsuruoka, Ryokan Ri and Jing Zhou for valuable discussions. Qiyu Wu was supported by JST SPRING, Grant Number JPMJSP2108.

\clearpage
\bibliographystyle{acl_natbib}
\bibliography{anthology,custom}

\clearpage
\appendix

\section{Augmentation Strategies}
\label{ap:implementation}
We propose diverse augmentation strategies for each sentence. In this paper we utilize five unsupervised augmentation strategies that are commonly adopted in previous works~\cite{Wei2019EDAED, yan2021consert, gao2021simcse}. Augmentations from discrete perspectives $\Delta^{(d)}$ includes:
1) Shuffled Sentence (SS) shuffles the position of words in the sentence. SS corrupts the order of the original sentence but preserves the semantic information of words;
2) Inverted Sentence (IS) inverts the original sentence as the augmented sample. Apart from the reading order, IS preserves all language properties even including n-gram statistics \citep{dufter2020identifying};
3) Words Repetition (WR) duplicates part of words and randomly insert them into the original sentences;
4) Words Deletion (WD) deletes part of words in the sentences.
WD and WR change the length and words of the original sentence but roughly preserve the reading order. The deletion and repetition ratio are empirically set to 0.2.
The augmentations from the continuous perspective $\Delta^{(c)}$ include Dropout (DP). It generates augmentation instances in the embedding level by passing the original sentence again into the encoder with different dropout masks.
The above five strategies can be repeatedly applied in practice. As there is randomness in the processes of augmentation and encoding, repeatedly generated instances with the same strategy can be regarded as diverse positives. But the diversity may accordingly decline.
Note that the primary goal of this paper is to verify the effectiveness of our PCL framework, hence all of the chosen augmentations are common and simple. We speculate that our PCL can be further improved with more fine-tuned augmentation strategies.

\begin{table*}[]
\centering
\small
\resizebox{1\textwidth}{!}{
\begin{tabularx}{\textwidth}{lXXXXXXXX}
\toprule
Model                    & MR             & CR             & SUBJ           & MPQA           & SST            & TREC           & MRPC           & Avg.           \\
\midrule
\multicolumn{9}{c}{Models w/o PLMs} \\
\midrule
GloVe embeddings (avg.)  & 77.25          & 78.30          & 91.17          & 87.85          & 80.18          & 83.00          & 72.87          & 81.52          \\
Skip-thought             & 76.50          & 80.10          & 93.60          & 87.10          & 82.00          & 92.20          & 73.00          & 83.50          \\
\midrule
\multicolumn{9}{c}{Base Models} \\
\midrule
Avg. BERT embeddings     & 78.66          & 86.25          & 94.37          & 88.66          & 84.40          &92.80 & 69.54          & 84.94          \\
BERT-\texttt{[CLS]} embedding & 78.58          & 84.85          & 94.21          & 88.23          & 84.13          & 91.40          & 71.13          & 84.66          \\
IS-BERT$_\text{base}$              & 81.09          & 87.18          & 94.96 & 88.75          & 85.96          & 88.64          & 74.24          & 85.83 \\
SimCSE-BERT$_\text{base}$          & 81.18          & 86.46          & 94.45          & 88.88          & 85.50          & 89.80          & 74.43          & 85.81          \\
SimCSE-RoBERTa$_\text{base}$       & 81.04          & 87.74 & 93.28          & 86.94          & 86.60          & 84.60          & 73.68          & 84.84          \\
\midrule
Ours-BERT$_\text{base}$           & 80.11          & 85.25          & 94.22          & 89.15 & 85.12          & 87.40          & 76.12          & 85.34          \\
Ours-RoBERTa$_\text{base}$         & 81.83 & 87.55          & 92.92          & 87.21          & 87.26 & 85.20          & 76.46 & 85.49          \\
\midrule
\multicolumn{9}{c}{Large Models} \\
\midrule
SimCSE-RoBERTa$_\text{large}$      & 82.74          & 87.87          & 93.66          & 88.22          & 88.58          & 92.00          & 69.68          & 86.11          \\
\midrule
Ours-BERT$_\text{large}$          & 82.47          & 87.87          & 95.04 & 89.59 & 87.75          & 93.00          & 76.00 & 87.39          \\
Ours-RoBERTa$_\text{large}$        & 84.47 & 89.06 & 94.60          & 89.26          & 89.02 & 94.20 & 74.96          & 87.94 \\
\bottomrule
\end{tabularx}
}
\caption{Transfer task results (measured as accuracy).}
\label{tab:transfer}
\end{table*}
\section{Baselines}
\label{ap:baselines}
We compare PCL with previous state-of-the-art unsupervised sentence embedding approaches. Basic approaches include taking the average of GloVe, BERT, or RoBERTa embeddings. Besides, BERT-whitening and BERT-flow post-process the embeddings distribution of BERT.
We also compare PCL with recent approaches using contrastive learning, including IS-BERT, ConSERT, SG-OPT, Contrastive Tension, DeCLUTR, and SimCSE. The following are the details of these baselines,
\begin{itemize}
\item GloVe \citep{pennington2014glove} maps words into a meaningful space where the distance between words is related to semantic similarity. The results of the average of GloVe embeddings are from \citet{reimers2019sentence}.  
\item \citet{su2021whitening} takes the average of the first and last layers of BERT~\cite{Devlin2019BERTPO} or RoBERTa~\cite{Liu2019RoBERTaAR} embeddings. We report the results from \citet{gao2021simcse}.
\item BERT-whitening~\cite{su2021whitening} and BERT-flow~\cite{Li2020OnTS} post-process the embeddings distribution of BERT. We report the results from \citet{gao2021simcse} for a fair comparison.
\item IS-BERT~\cite{Zhang2020AnUS} encourages the representation of a specific sentence to encode all aspects of its local context information, using local contexts derived from other input sentences as negative examples for contrastive learning. We report the results from the original paper.
\item ConSERT~\cite{yan2021consert} contrasts a pair of sentences augmented by different augmentation methods. We report the results in the original paper.
\item SG-OPT~\cite{Kim2021SelfGuidedCL} is a contrastive learning method using self-guidance. The results are from the original paper.
\item Contrastive Tension (CT)~\cite{Carlsson2021SemanticRW} propose a training objective that aligns the embeddings of the same sentence encoded by two different encoders. We report the results from \citet{gao2021simcse}.
\item DeCLUTR~\cite{Giorgi2021DeCLUTRDC} is a contrastive approach that takes different spans from the same document as contrastive pairs. The results are from \citet{gao2021simcse}
\item Mirror-BERT~\cite{liu2021fast} employs several fast augmentation strategies for effective representations. The results are from the original paper.
\item SimCSE~\cite{gao2021simcse} contrasts a pair of embeddings of one sentence encoded with different dropout masks. The results are from the original paper. We had re-run SimCSE with same setups and it performs worse than the numbers reported in the original paper (e.g., 75.36 averaged over 5 seeds on BERT$_\text{base}$). We reported higher numbers for a fair comparison.
\end{itemize}
Due to the surge of this topic, many concurrent works emerge with two trends: \textit{new model structure} and \textit{more in-domain data}.
SGPT~\cite{muennighoff2022sgpt} and Sentence-T5~\cite{ni2021sentence} are proposed with new paradigm and far larger models, which underperform with comparable model size. Trans-Encoder~\cite{liu2021trans} proposes a cooperative method with in-domain pairwise data for mutual benefits of bi- and cross-encoder, making the results incomparable.
\begin{table*}[]
    \begin{center}
    \centering
    \small
    \resizebox{1\textwidth}{!}{
    \begin{tabularx}{\textwidth}{lYYYYYYYY}
    \toprule
       \textbf{Model} & \textbf{STS12} & \textbf{STS13} & \textbf{STS14} & \textbf{STS15} & \textbf{STS16} & \textbf{STSb} & \textbf{SICK-R} & \textbf{Avg.} \\
    \midrule
    \multicolumn{9}{c}{Effect of the number of augmentations} \\
    \midrule
    PCL$_\text{K=1}$ & 72.58 & 80.62 & 74.15 & 82.31 & 79.23 & 78.47 & 72.06 & 77.06 \\
    PCL$_\text{K=3}$ & 72.66 & 82.96 & 74.44 & 81.94 & 78.38 & 77.93 & 71.55 & 77.12 \\
    PCL$_\text{K=5}$ & 73.44 & 81.80 & 74.59 & 82.63 & 79.40 & 79.05 & 72.25 & 77.59 \\
    PCL$_\text{K=7}$ & 73.49 & 81.93 & 74.84 & 82.24 & 79.75 & 79.37 & 72.62 & 77.75 \\
    PCL$_\text{K=9}$ & 72.74 & 83.36 &	76.05 & 83.07 & 79.26 & 79.72 & 72.75 & 78.14 \\
    \midrule
    \multicolumn{9}{c}{Effect of the diversity of augmentations} \\
    \midrule
    PCL$_\text{DP}$ & 71.20 & 82.53 & 74.66 & 82.67 & 78.92 & 78.06 & 71.94 & 77.14 \\
PCL$_\text{SS}$ & 70.60 & 80.73 & 74.11 & 82.18 & 78.90 & 77.91 & 69.69 & 76.30 \\
PCL$_\text{IS}$ & 70.95 & 81.31 & 74.51 & 82.24 & 79.23 & 78.44 & 72.09 & 76.97 \\
PCL$_\text{WR}$ & 71.82 & 82.56 & 74.75 & 82.34 & 78.85 & 78.72 & 71.88 & 77.27 \\
PCL$_\text{WD}$ & 73.08 & 81.84 & 74.17 & 82.50 & 78.81 & 78.52 & 71.23 & 77.17 \\
    PCL & 72.74 & 83.36 &	76.05 & 83.07 & 79.26 & 79.72 & 72.75 & 78.14 \\
    \bottomrule
    \end{tabularx}
    }
    \end{center}
    \caption{
        Effect of the number and diversity of augmentations.
        We report the Spearman's correlation $\rho$ (\%) on 7 STS datasets. All variants are run for five times with different random seeds and the \textit{average} of these five results on each column is reported as the final number.
    }
    \label{tab:ablation_k}
\end{table*}

\section{Additional Experimental results}
\subsection{Comparison of controlled setups}
\label{ap:control}
We can also find some variants of SimCSE in our controlled experiments. For example, PCL$_{noP}$ in Table \ref{tab:ablation} is regarded as \emph{SimCSE w/ multi-augmentations}. PCL$_{K=1}$ in Figure \ref{fig:ablation_k} can be regarded as \textit{ SimCSE w/ peer-network cooperation}, and PCL$_{DP}$ in Figure~\ref{fig:ablation_diversity} is regarded as \emph{SimCSE w/ 9 dropout augmented positive samples}. As we discussed in the \S~\ref{sec:ablation_augs} and \S~\ref{sec:ablation}, the comparison between the variants of SimCSE and PCL show the advantages and importance of our proposed peer-contrast and peer-cooperation.

\subsection{Transfer tasks}
\label{ap:transfer}
We also evaluate PCL on 7 transfer tasks~\cite{conneau2018senteval}. As the Table~\ref{tab:transfer} shows, PCL achieves competitive performance compared with baselines. Note that as mentioned in previous works~\cite{reimers2019sentence,gao2021simcse}, the main goal of sentence embeddings is to cluster semantically similar sentences. Hence we only take STS as the main results in this paper.

\subsection{Detailed experimental results on analysis of diverse augmentations}
\label{ap:diverse}
In this section, we present detailed results of experiments of diverse augmentations on all 7 STS tasks. We test the performance of PCL with varying $K$ and diversity. Experimental results are shown in Table~\ref{tab:ablation_k}. As the results and our analysis in \S~\ref{sec:ablation_augs} show, the performance of PCL maintains an upward trend with increasing $K$. Besides, it is also shown that PCL significantly outperforms its mono-augmenting variants, even keeping the $K$ of them constant, which indicates a better generalization. As a result, PCL with more diverse augmentations performs better. We also speculate that our PCL can be further improved with larger $K$ and more fine-tuned augmentation strategies.

\section{Discussion}
\subsection{Distinction between PCL and other contemporary methods.}
\label{ap:realted}
\paragraph{Unsupervised sentence embedding w/ multiple positive augmentations.} 
It's straightforward to extend mono-augmentation into multi-augmentation to learn expressive representations. For example, CLEAR~\cite{wu2020clear} uses various token/span manipulations for noise-invariant representations while Mirror-BERT~\cite{liu2021fast} employs several fast augmentation strategies for effective representations. However, these methods usually take the augmented positives equally, regardless of the \textbf{uncontrolable qualities}. 
For example, given ``\textit{Two men are wrestling on the floor}'', we get ``\textit{Two men are squirming on the floor}'' and ``\textit{Two persons are wrestling on the floor}'' by word replacement, but only the 2nd is reasonable. 
Thereby, we take a step further to consider the contrasts among the augmented positives to figure out which augmentation is relatively reasonable. This is achieved by our novel cooperative learning method with peer networks. 

\paragraph{Unsupervised sentence embedding for anti-bias.} 
More related to our work, ESimCSE~\cite{Wu2021ESimCSEES} found learning on dual-dropout causes sentence length bias so it employs another augmentation strategy, i.e., word repetition, to prevent the length bias. However, word repetition introduces \textbf{learning shortcut} (i.e., \textit{order} and \textit{BoW} as in Table \ref{tab:bias}) by itself, not to mention it makes the sentence unnatural and even \textbf{semantics-wrong} (e.g., repeating ``\textit{no}''). 
To circumvent this dilemma, we propose exhaustive augmentations to ensure ``\textit{at-least-one}'' true positive and reduce learning shortcuts by complementary augmenting strategies 
(See Figure \ref{fig:intro} and \ref{fig:ablation_diversity}: if we employ every strategy, the shortcuts can be blocked).
Nonetheless, PCL still has a better performance (78.42 vs. 78.27) compared with ESimCSE on STS tasks.

\paragraph{Cooperative learning has more parameters, is it the reason leading better performance?}
One of the advantages of our cooperative learning is to effectively learn from the positives with uncontrolled qualities, or it can be also interpreted as `noisy labels'. In the noisy circumstance, more parameters not necessarily lead to better performance. And it can be anticipated that it possibly leads to worse performance because the over fitting in the noisy positives.

\subsection{Efficiency \& Impact of augmentations.}
\paragraph{Efficiency} Due to peer positives and encoders, our framework needs higher (\textasciitilde3$\times$, i.e., 2.5 GPU-hours for base models) computation overheads compared to vanilla CL \cite{gao2021simcse}. Nonetheless, the same inference makes our framework still practical and scalable. Since PCL contains more loss items, we do not see any significant difference in convergence time.
\paragraph{Impact of augmentations} The performance of our framework relies on the augmentation methods, and it is hard to claim which combination of the methods is optimal except experimental verification. In this work, we only intuitively select several methods without extensive trials. We have illustrated the performance of mono-augmentation in Figure \ref{fig:ablation_diversity}.



\end{document}